%% file: main.tex
\documentclass[runningheads]{llncs}
\usepackage{lipsum}
\usepackage[T1]{fontenc}
\usepackage{multirow}
\usepackage{graphicx}
\usepackage{color}
\usepackage{amsmath}
\usepackage{amsfonts}
\usepackage{xcolor}
\usepackage{graphicx}
\usepackage{subcaption}
\usepackage{caption}
\usepackage{float}

\usepackage{afterpage}
\usepackage{placeins}
\usepackage{enumitem}
\usepackage{comment}
\usepackage{arydshln}
\usepackage{ulem}
\usepackage{hyperref}
\usepackage{pdfpages}

\setlist{topsep=0pt, partopsep=0pt, parsep=0pt, itemsep=5pt}

\begin{document}

\title{Offline Handwritten Signature Verification Using a Stream-Based Approach}
\author{Kecia G. de Moura\inst{}
\and
Rafael M. O. Cruz\inst{}
\and
Robert Sabourin\inst{}
}

\authorrunning{K. G. de Moura et al.}
\institute{École de technologie supérieure - Université du Québec, Montreal, Québec, Canada
\email{kecia.gomes-de-moura.1@ens.etsmtl.ca}\\
\email{\{rafael.menelau-cruz,robert.sabourin\}@etsmtl.ca}
}
\maketitle 
\vspace{-10pt}
\begin{abstract}
Handwritten Signature Verification (HSV) systems distinguish between genuine and forged signatures. Traditional HSV development involves a static batch configuration, constraining the system's ability to model signatures to the limited data available. Signatures exhibit high intra-class variability and are sensitive to various factors, including time and external influences, imparting them a dynamic nature. This paper investigates the signature learning process within a data stream context. We propose a novel HSV approach with an adaptive system that receives an infinite sequence of signatures and is updated over time. Experiments were carried out on GPDS Synthetic, CEDAR, and MCYT datasets. Results demonstrate the superior performance of the proposed method compared to standard approaches that use a Support Vector Machine as a classifier. Implementation of the method is available at \url{https://github.com/kdMoura/stream_hsv}.

\end{abstract}

\input{sections/0_sections}

 \bibliographystyle{splncs04}
 \bibliography{references}

 

\section*{Supplementary material}

# Offline Handwritten Signature Verification Using a Stream-Based Approach
# – Supplementary material –

Kecia G. de Moura, Rafael M. O. Cruz, and Robert Sabourin

École de technologie supérieure - Université du Québec, Montreal, Québec, Canada
kecia.gomes-de-moura.1@ens.etsmtl.ca
{rafael.menelau-cruz,robert.sabourin}@etsmtl.ca

## Overview

This supplementary material provides experimental results that complement the main paper.

Table 1: Skilled forgery detection in batch settings for different development sets using SVM and SGD. $n\mathcal{D}$ and #G denote, respectively, the number of users and genuine signatures per user ($n\mathcal{D}_G$) employed during training.

| $n\mathcal{D}$ | $n\mathcal{E}_R$ | #G sig. = 2 | | #G sig. = 6 | | #G sig. = 12 | |
|---|---|---|---|---|---|---|---|
| | | SVM | SGD | SVM | SGD | SVM | SGD |
| 2 | 1 | 19.82 (1.01) | 61.97 (8.92) | 19.39 (1.00) | 68.52 (3.94) | 19.80 (0.47) | 60.93 (7.80) |
| | 2 | 16.96 (0.95) | 62.15 (8.70) | 16.84 (0.54) | 68.11 (3.52) | 17.15 (0.47) | 61.17 (8.03) |
| | 3 | 15.57 (0.95) | 62.53 (8.06) | 15.48 (0.72) | 68.31 (3.33) | 15.59 (0.37) | 61.61 (7.89) |
| | 5 | 14.17 (0.86) | 62.64 (7.70) | 14.03 (0.44) | 68.01 (3.56) | 14.34 (0.48) | 61.74 (7.25) |
| | 10 | 13.07 (1.02) | 62.63 (7.37) | 12.83 (0.38) | 67.88 (3.30) | 13.11 (0.40) | 61.84 (6.81) |
| | 12 | 12.79 (0.92) | 62.54 (7.34) | 12.41 (0.40) | 67.73 (3.27) | 12.72 (0.44) | 62.10 (6.93) |
| 5 | 1 | 19.02 (0.54) | 57.47 (3.87) | 18.37 (0.75) | 47.26 (7.88) | 18.25 (0.66) | 23.60 (1.81) |
| | 2 | 16.23 (0.21) | 58.23 (3.37) | 15.79 (0.21) | 47.87 (8.49) | 15.57 (0.10) | 20.86 (1.91) |
| | 3 | 14.87 (0.44) | 58.68 (3.40) | 14.07 (0.36) | 48.02 (9.17) | 14.31 (0.27) | 19.37 (1.82) |
| | 5 | 13.51 (0.62) | 58.99 (3.10) | 12.71 (0.34) | 48.47 (9.36) | 12.79 (0.43) | 18.19 (1.84) |
| | 10 | 12.23 (0.44) | 59.15 (3.45) | 11.77 (0.27) | 48.84 (9.66) | 11.75 (0.30) | 16.91 (1.88) |
| | 12 | 11.92 (0.29) | 59.17 (3.14) | 11.43 (0.17) | 49.05 (9.85) | 11.59 (0.21) | 16.63 (1.77) |
| 10 | 1 | 18.12 (1.01) | 37.53 (3.38) | 17.14 (0.97) | 19.02 (1.09) | 17.03 (0.96) | 17.54 (1.04) |
| | 2 | 15.58 (0.57) | 36.70 (3.99) | 14.22 (0.39) | 16.01 (0.54) | 14.11 (0.34) | 14.69 (0.25) |
| | 3 | 14.14 (0.42) | 36.38 (4.29) | 12.86 (0.49) | 14.44 (0.70) | 12.63 (0.17) | 13.19 (0.24) |
| | 5 | 12.75 (0.55) | 36.56 (4.01) | 11.51 (0.42) | 13.07 (0.57) | 11.34 (0.41) | 11.78 (0.31) |
| | 10 | 11.47 (0.47) | 36.98 (4.42) | 10.39 (0.54) | 11.95 (0.86) | 10.22 (0.46) | 10.58 (0.49) |
| | 12 | 11.20 (0.35) | 37.05 (4.83) | 10.17 (0.59) | 11.65 (0.67) | 9.99 (0.49) | 10.65 (0.49) |
| All | 1 | 16.14 (0.69) | 16.81 (0.78) | 15.17 (0.50) | 15.25 (0.44) | 15.08 (0.62) | 14.93 (0.65) |
| | 2 | 13.50 (0.54) | 14.01 (0.39) | 12.36 (0.13) | 12.59 (0.24) | 12.28 (0.56) | 12.23 (0.53) |
| | 3 | 12.12 (0.16) | 12.69 (0.51) | 11.13 (0.13) | 11.21 (0.26) | 11.04 (0.29) | 11.09 (0.29) |
| | 5 | 10.71 (0.20) | 11.34 (0.36) | 9.92 (0.34) | 10.13 (0.13) | 9.71 (0.27) | 9.86 (0.28) |
| | 10 | 9.59 (0.42) | 10.33 (0.43) | 8.89 (0.36) | 9.17 (0.39) | 8.75 (0.44) | 8.89 (0.44) |
| | 12 | 9.42 (0.48) | 10.04 (0.36) | 8.72 (0.43) | 9.04 (0.36) | 8.46 (0.25) | 8.72 (0.36) |

Table 1 presents the results of skilled forgery detection in batch settings using Support Vector Machine (SVM) and Stochastic Gradient Descent (SGD) classifiers. Results show the Equal Error Rate (EER) with global thresholds for a writer-independent approach. Various configurations for the development set $\mathcal{D}$, involving different numbers of users $n\mathcal{D}$ and genuine signatures per user $n\mathcal{D}_G$, were employed, resulting in different trained models for SVM and SGD. The trained classifiers were then applied to the exploitation set $n\mathcal{E}$ (GPDS-S-300) with varying numbers of reference signatures $n\mathcal{E}_R$. The results indicate a clear relationship: as the number of users and the volume of signatures per user during the training phase increase, so does the performance of the classifiers. In summary, higher quantities of users and signatures lead to better detection.

In [1], SVM was trained using a development set with 2000 users. The classier was applied on a GPDS-S-300 exploitation set, obtaining an EER of $7.93 \pm 0.30$ using 12 reference signatures. Although SGD in batch configurations has poor performance, the proposed SHSV is able to reach comparable results to the ones reported in [1] even starting with a limited development set.

Table 2: EER on last and average $\mathbb{T}_j$ chunks of signatures stream. Results show skilled forgery detection on signatures randomly coming from the CEDAR and MCYT streams using SVM and the Stream HSV (SHSV). $n\mathcal{D}$ refers to the number of users in the training set, and $n\mathcal{D}_G$ is the number of genuine signatures per user. $n\mathcal{E}_R$ denotes the number of reference signatures in the test set. For the stream evaluation: test $c_{size} = 300$, $w_{size} = w_{step} = 200$ signatures.

| $n\mathcal{D}$ | $n\mathcal{E}_R$ | #G sig. ($n\mathcal{D}_G$) = 12 | | | | | |
| --- | --- | --- | --- | --- | --- | --- | --- |
| | | $SVM_{last}$ | $SHSV_{last}$ | $\Delta$ | $SVM_{avg}$ | $SHSV_{avg}$ | $\Delta$ |
| 2 | 1 | 20.60 (0.89) | 19.80 (2.77) | 0.80 | 20.70 (2.31) | 23.21 (3.28) | 2.51 |
| | 2 | 18.00 (1.41) | 17.60 (2.41) | 0.40 | 18.43 (2.34) | 21.61 (3.22) | 3.18 |
| | 5 | 16.60 (2.51) | 14.40 (2.79) | 2.20 | 16.36 (1.80) | 20.25 (3.07) | 3.89 |
| | 10 | 15.20 (2.59) | 13.20 (1.92) | 2.00 | 14.92 (2.15) | 18.65 (2.60) | 3.73 |
| 5 | 1 | 21.60 (1.52) | 18.60 (2.97) | 3.00 | 21.17 (2.44) | 18.92 (2.39) | 2.25 |
| | 2 | 19.40 (1.52) | 17.00 (2.12) | 2.40 | 19.57 (2.31) | 16.70 (2.29) | 2.87 |
| | 5 | 17.20 (2.39) | 14.00 (2.35) | 3.20 | 17.39 (2.09) | 14.90 (2.18) | 2.49 |
| | 10 | 16.40 (2.51) | 11.40 (3.05) | 5.00 | 16.05 (2.03) | 13.32 (2.41) | 2.74 |
| 10 | 1 | 20.20 (1.64) | 18.60 (3.05) | 1.60 | 20.59 (2.17) | 18.79 (2.51) | 1.80 |
| | 2 | 18.60 (0.89) | 17.80 (3.03) | 0.80 | 18.79 (2.21) | 16.79 (2.31) | 2.00 |
| | 5 | 16.00 (2.55) | 14.40 (2.07) | 1.60 | 16.47 (2.22) | 14.61 (2.31) | 1.87 |
| | 10 | 13.80 (2.39) | 12.20 (3.27) | 1.60 | 14.79 (1.65) | 12.85 (2.53) | 1.93 |
| All | 1 | 20.00 (1.00) | 18.40 (2.70) | 1.60 | 19.94 (2.31) | 18.90 (2.59) | 1.05 |
| | 2 | 19.20 (1.64) | 17.20 (1.92) | 2.00 | 18.50 (2.38) | 16.67 (2.29) | 1.82 |
| | 5 | 16.40 (2.07) | 14.60 (1.67) | 1.80 | 16.63 (2.17) | 14.56 (2.13) | 2.07 |
| | 10 | 15.60 (2.70) | 12.40 (2.51) | 3.20 | 14.94 (2.13) | 13.27 (2.36) | 1.67 |

Table 2 reports the Equal Error Rate of the pre-trained models (using different development sets) on a stream of signatures from an unknown distribution. While the models were trained on GPDS Synthetic, the arriving test signatures belonged to CEDAR or MCYT datasets. To this, the stream generation protocol was applied to the CEDAR and MCYT exploitation sets. The two streams were combined by randomly selecting signatures from each one. For the model update, we employed a test chunk size ($c_{size}$) of 300 signatures, consisting of 100 genuine signatures, 100 random forgeries, and 100 skilled forgeries, resulting in a balanced chunk: $Chunk = \{S_{G_1}, S_{RF_1}, S_{SK_1}, S_{G_2}, S_{RF_2}, S_{SK_2}, ..., S_{G_{300}}, S_{RF_{300}}, S_{SK_{300}}\}$. In this setup, 300 signatures are tested, and 200 signatures (excluding the skilled forgeries) are used to update the classifier, continuously performing the test-then-train procedure until the stream's end. Each chunk was evaluated regarding EER, considering the results from genuine and skilled forgeries.

Results in Table 2 show the performance of models on the last and average chunks of the stream. Although SHSV is impacted by the initial learning, we see that it can outperform the SVM in most configurations, delivering improved verification of signatures over time.


\end{document}

%% file: sections/0_sections.tex
\input{sections/1_introduction}

\input{sections/3_proposed_method}

\input{sections/4.1_experimental_setup}

\input{sections/4.2_experimental_results}

\input{sections/5_conclusion}


%% file: sections/1_introduction.tex
\section{Introduction} 

Handwritten Signature Verification (HSV) systems aim to automatically distinguish between genuine signatures, belonging to the claimed individual, and forgeries. In offline HSV, signatures are represented as digital images captured after the writing process is completed, as opposed to online systems that analyze the signing dynamics \cite{Hameed2021}.

Offline HSV systems can be categorized into two approaches: writer-dependent (WD) and writer-independent (WI). In WD systems, a unique classifier is trained for each enrolled user, offering potentially higher accuracy. However, this approach requires individual training data for each new user. Conversely, WI systems utilize a single classifier for all users, hence being more scalable \cite{Hafemann2017_literatureReview}. In this case, the classification is performed on a dissimilarity space where the pattern recognition is reduced to a 2-class problem by using differences between claimed and reference signatures through a
Dichotomy Transformation (DT) \cite{Hafemann2017_literatureReview}.

In general, HSV development entails two distinct datasets: a development set utilized for training and an exploitation set employed during the testing phase \cite{Hafemann2017_learning}. Each set comprises the signatures of enrolled users. While more samples generally lead to better generalization models, real-world applications often face data availability limitations regarding both user count and sample volume \cite{Hafemann2017_literatureReview}. Therefore, the system's capability to model signature variations is constrained to the present available data. After the training process, the resulting HSV is expected to achieve generalization to the whole set of existing users and their signatures. 

Nonetheless, by relying on a training process with a finite dataset, the current literature does not account for the inherent variability and changing behavior of handwriting signatures. Signatures exhibit the highest intra-class variability compared to other biometric traits \cite{Hafemann2017_literatureReview}. Additionally, signature patterns are time-sensitive as they evolve as we age. Besides, diverse factors can impact the signing process, including emotional states, stress levels, fatigue, and influences from substances like alcohol or drugs \cite{Diaz2019_perspective}. Writing results are intrinsically related to cognitive-motor and neuromotor conditions, being affected by any minor impairment \cite{Diaz2019_perspective}.

Given these challenges, we pose the following general
research question: \textit{how can a signature verification system adapt to the inherent variability and evolving nature of handwritten signatures over time, maintaining high verification performance while mitigating the problem of limited data?} To answer this question, we propose a framework to handle signature verification in an adaptive manner, where the input data is processed as a stream of offline signatures rather than a batch mode. 

In the proposed framework, incoming signatures are first tested and then used to improve the system by updating its current state. In this approach, \textit{SigNet-S} \cite{Talles2023_AmultitaskApproach4ContrastiveLearning}, one of the state-of-the-art representation models, is employed to extract features of incoming claimed signatures. These feature vectors are then compared to corresponding reference vectors stored in the database to create dissimilarity samples via a stream dichotomy transformation. Lastly, the adaptive WI-classifier is updated based on the dissimilarity vectors. To the best of our knowledge, no prior work has considered signatures in an open-set, stream-based configuration.

The main contributions of this article are as follows:
\begin{enumerate}
    \item Stream HSV: we propose a novel HSV framework that adapts over time. This framework treats signatures as an infinite data stream, enabling continuous learning and improvement.
    \item A stream dichotomy transformation: we introduce a stream dichotomy transformation process to facilitate adaptive learning from the incoming signature stream and address the challenge of imbalanced data ratios commonly encountered in stream scenarios.
    \item Signature stream generation method: to facilitate evaluation using standard batch configurations, we introduce a method for generating signature streams based on the existing HSV evaluation protocol.
\end{enumerate}

%% file: sections/3_proposed_method.tex
\section{Stream handwriting signature verification (SHSV)}
\label{sec:proposed}
\vspace{-2pt}
Streaming systems are designed to continuously process and analyze data as it arrives, delivering updated results based on the most recent characteristics of the information. In the context of handwriting signature verification, this approach enables adaptive signature authentication, accounting for the inherent variability of this biometric modality. As the system evolves to accommodate variations in handwriting patterns, it ensures accurate and reliable verification.

In light of this, we propose a framework for signature verification under a data stream context. Specifically, we introduce a system that takes as input a sequence of signatures $S$ of claimed users $\hat{l}$, denoted as $Stream = \{(S, \hat{l})_{1}, (S, \hat{l})_{2}, ...,  \allowbreak (S, \hat{l})_{\infty}\}$,  for verification against the signatures of enrolled users. As new signature samples arrive (from new or enrolled users), the system incorporates new information into its base knowledge and delivers updated results on the next verification. The system is depicted in Figure~\ref{fig:fig_onlineASV}, and notation is synthesized in Table~\ref{tab:symbols}.

\input{tables/tab_symbols}

\input{figures/fig_onlineASV/main}

The SHSV is a general framework that comprises fundamental components that enable the system to work in a writer-independent (WI) manner. They are following described.

\subsection{Representation model $\phi(\cdot)$}
The representation model is a previously well-trained model capable of extracting relevant features from the signature images. To this end, the SHSV employs the SigNet Synthetic (\textit{SigNet-S}) developed by \cite{Talles2023_AmultitaskApproach4ContrastiveLearning}. This model is a variant of the original \textit{SigNet} proposed by \cite{Hafemann2017_learning}. While the original \textit{SigNet} was trained using signature examples obtained from the GPDS-960 Grayscale \cite{Vargas2007_DB_GPDS960}, which is no longer publicly available due to the General Data Protection Regulation (EU) 2016/679, \textit{SigNet-S} was trained using synthetic GPDS data \cite{Ferrer2015_DB_GPDSsynthetic}.

\textit{SigNet-S} employs Deep Convolutional Neural Networks (DCNNs) to learn signature features, focusing on capturing key attributes that differentiate individual writers. The model adopts a writer-independent training approach, utilizing only genuine signatures during training, which allows for adaptability to new users. When extracting features from the signatures of new users, the network performs feed-forward propagation up to the fully connected layer preceding the Softmax layer, producing feature vectors with a dimensionality of 2048. In this framework, these vectors define the feature space for each signature. Formally, given a signature image $S_C$ from a claimed user, the corresponding feature vector is represented as $\mathbf{x}_C = \phi(S_C)$.

\subsection{Stream dichotomy transformation $DT(,)$}
An essential part of the  SHSV system is the dichotomy transformation $DT(,)$ \cite{Eskander2013_DissimilarityRepresentation}. It transforms a multi-class problem into a 2-class problem. This enables the implementation of a writer-independent approach for the classification task, which is crucial for the stream context. The binary-problem result is achieved by computing the absolute distance between each feature of two feature vectors, i.e., the dissimilarity between two samples. 
Suppose $(\mathbf{x}_R, l_R)$ and $(\mathbf{x}_C, l_C)$ are the feature vector ($\mathbf{x}$) and label ($l$) of two data samples, where $l$ refers to the author's ID. With $\mathbf{x}_R = \{f_{k}^{R}\}_{k=1}^{K}$ and $\mathbf{x}_C = \{f_{k}^{C}\}_{k=1}^{K}$, where $K$ is the number of features $f$. The dissimilarity vector between $\mathbf{x}_R$ and $\mathbf{x}_C$ is given by $\mathbf{\dot{x}}_{RC}=DT(\mathbf{x}_R,\mathbf{x}_C) = \{|f_{k}^{R}- f_{k}^{C}|\}_{k=1}^{K}$, where $\left| \: \cdot \: \right|$ represents the absolute value of the difference. The vector $\mathbf{\dot{x}}_{RC}$ has the same dimensionality as $\mathbf{x}_R$ and $\mathbf{x}_C$. 

The resulting dissimilarity set after applying $DT(,)$ on $(\mathbf{x}_R, l_R)$ and $(\mathbf{x}_C, l_C)$ is given by $(\mathbf{\dot{x}}_{RC}, y_{RC})$ where $y$ denotes the new label. If $l_R = l_C$, i.e., $\mathbf{\dot{x}}_{RC}$ is obtained from signatures of the same writer, it is labeled as \textit{positive} ($y = +$). Otherwise, it will be labeled as \textit{negative} ($y = -$), i.e., $l_R \neq l_C$. When the claimed and the reference signature are similar, the corresponding dissimilarity vector is expected to be located near the origin. In contrast, the negative samples are expected to have a sparse distribution in space \cite{Souza2019_CharacterizationOfHSinDissimilarityRepresentationSpace}. 

In this work, $DT(,)$ is applied to the streaming of claimed signatures against each correspondent reference sample stored in the database of enrolled users. Specifically, consider $\mathbb{S}^{i}_{R} = \{S^{i,1}_{R}, S^{i,2}_{R}, \ldots ,S^{i,M}_{R}\}$  the set of $M$ reference signatures of user $i$, and $S_C$ a claimed signature of same user.  Then,  $DT(\phi(\mathbb{S}^{i}_{R}), \phi(S_C)))$ results in the correspondent dissimilarity set $\mathbb{\dot{X}}^{i}_{RC} = \{\mathbf{\dot{x}}^{i,1}_{RC}, \allowbreak \mathbf{\dot{x}}^{i,2}_{RC}, \ldots  ,\mathbf{\dot{x}}^{i,M}_{RC}\}$, which is passed to the adaptive WI-classifier for training and testing procedure. 

 The dichotomy transformation also helps mitigate the common imbalance ratio issues in streaming data. For each incoming genuine signature, it is always possible to generate the same amount of negative dissimilarities by utilizing the user's stored reference signatures and selecting the necessary number of random forgery samples. This approach consistently produces an equal number of positive and negative examples.  

\subsection{Adaptive WI-classifier $\theta$}

In the proposed SHSV approach, the core component is the adaptive verification process, which enables the system to update its base knowledge over time. In static HSV systems, Support Vector Machines (SVM) are a popular choice for the verification step \cite{Hameed2021}.  Nonetheless, an adaptive classifier is required for the present work. Many methods have been developed to adapt the traditional SVM to handle evolving data \cite{Zhou_2016_onlineSVM_survey}. An efficient optimization method is applying Stochastic Gradient Descent (SGD) to linear models to minimize the loss function \cite{Losing_2018_onlineSVM_IncrementalonlinelearningReview}. To mimic the SVM behavior with adaptive capability, we adopt the SGD classifier with a hinge loss function. We follow similar works\cite{Panagiotakopoulos_2013_SVMSGD_TheStochasticGradientDescentforthePrimal,ShalevShwartz_2011_SVMSGD_Pegasos,Zhai_2017_SVMSGD_EBPegasos_ClassificationHighdimensional} that employed SGD to minimize the loss function in the primal formulation directly. This approach is more efficient than employing Lagrangian methods as it avoids the need to compute and store dual variables, which can become computationally expensive and memory-intensive as the number of data points and features increases. The loss function, described in Equation~\ref{eq:svm_primal}, aims to minimize the norm of the weight vector 
$\mathbf{w}$ while penalizing misclassifications (quantified by the hinge loss term $\max(0, 1 - y_i (w \cdot \mathbf{x}_i + b))$), where $C$ is the regularization parameter that controls the trade-off between maximizing the margin and minimizing the hinge loss.

\begin{equation}
\label{eq:svm_primal}
\min_{\mathbf{w}, b} \quad \frac{C}{2} ||\mathbf{w}||^2 + \frac{1}{N} \sum_{i=1}^{N} \max(0, 1 - y_i (\mathbf{w} \cdot \mathbf{x}_i + b))
\end{equation}

SGD processes individual samples or small batches from the dataset, iteratively updating the model parameters based on the loss function. In SHSV, the WI-classifier is updated with dissimilarity vectors obtained from a chunk of incoming signatures. This update process occurs after the classifier's prediction on the input chunk. For the present work, we assume that all true labels are available immediately after the classifier's estimation.

\subsection{Fusion function $\mathtt{g}(\cdot)$}

In SHSV, the WI-classifier's output is determined by the dissimilarity vector's distance to its decision hyperplane. When there is a set of reference signatures $\mathbb{S}_R$ for a claimed signature $S_C$, the system delivers a distance for each pair of dissimilarity vectors between $\mathbb{S}_R$ and $S_C$. These hyper-plane distances are combined through a fusion function $\mathtt{g}(\cdot)$ \cite{Rivard2013_fusion_function}. Results in \cite{Souza2018_WICNN} reveal that better verification performance is achieved when the Max fusion function is chosen to combine hyper-plane distances output. This work employs the maximum distance to deliver a final decision.

%% file: tables/tab_symbols.tex
\begin{table}
    \caption{SHSV notation.}
    \label{tab:symbols}
    \centering
        \begin{tabular}{
       p{0.9cm}|p{4cm}|p{0.8cm}|p{6cm}
       }
            \hline
            \textbf{Sym.} & \textbf{Description} & \textbf{Sym.} & \textbf{Description} \\
            \hline

$S$	&	Signature image	&	$\mathbb{\dot{X}}$	&	Set of dissimilarity vectors	\\
$l$	&	Writer label	&	$\mathcal{D}$	&	Development set	\\
$\phi(\cdot)$	&	Feature extractor	&	$n\mathcal{D}$	&	Number of writers in $\mathcal{D}$	\\
$DT(,)$	&	Dichotomy transformation	&	$n\mathcal{D}_G$	&	Number of genuine sig. per writer in $\mathcal{D}$	\\
$\theta$	&	WI-classifier	&	$\mathcal{E}$	&	Exploitation set	\\
$\mathtt{g}(\cdot)$	&	Fusion function	&	$n\mathcal{E}$	&	Number of writers in $\mathcal{E}$	\\
$\mathbf{x}$	&	Feature vector $\mathbf{x} = \phi(S)$	&	$n\mathcal{E}_R$	&	Number of ref. sig. per writer in $\mathcal{E}$	\\
$\mathbf{\dot{x}}$	&	Dissimilarity vector $\mathbf{\dot{x}}$ = $DT(\mathbf{x_1},\mathbf{x_2})$	&	$n\mathcal{E}_C$	&	Number of G, RF, and SK claimed sig. per writer in $\mathcal{E}$	\\
$y$	&	Dissimilarity label	&	$\mathbb{T}$	&	Stream obtained from $\mathcal{E}$	\\
$G$	&	Genuine signature	&	$\mathbb{T}_k$	&	Chunk $k$ of arriving signatures from $\mathbb{T}$	\\
$RF$	&	Random forgery signature	&	$S^{i,j}_G$	&	$j$-th genuine signature of writer $i$	\\
$SK$	&	Skilled forgery signature	&	$\dot{x}^{i,k,j}_G$	&	Dissimilarity vector between reference $k$ and genuine signature $j$ of writer $i$	\\
$R$	&	Reference signature	&	$\dot{\mathbb{X}}^{i,j}_G$	&	Set of dissimilarities from all references and genuine signature $j$ of writer $i$	\\
$C$	&	Claimed signature	&	$c_{\text{size}}$	&	Chunk size for model update	\\
$\mathbb{S}$	&	Set of signatures	&	$w_{\text{size}}$	&	Window size for stream evaluation	\\
$\mathbb{X}$	&	Set of feature vectors	&	$w_{\text{step}}$	&	Step size for evaluation frequency	\\

            \hline
        \end{tabular}
    
\end{table}

%% file: figures/fig_onlineASV/main.tex
\begin{figure*}
    \centering
    \includegraphics[width=\linewidth]{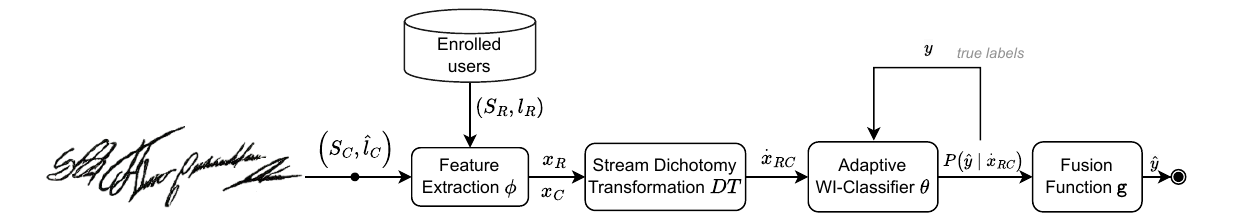}
    \caption{Stream HSV system. $(S_C,\hat{l}_C)$ denotes a claimed signature from the stream, and $(S_R,l_R)$ a reference signature of the corresponding user. Signatures, after being preprocessed, have their features extracted by a representation model $\phi$. The stream dichotomy transformation is applied to the pair of features vectors $DT(\textbf{x}_R, \textbf{x}_C)$ and passed to the adaptive classifier $\theta$, which outputs a prediction. At the end, a fusion function is employed considering all reference signatures to deliver a final result. If true labels are available, the classifier is updated with all new dissimilarities information. }
    \label{fig:fig_onlineASV}
\end{figure*}

%% file: sections/4.1_experimental_setup.tex
\section{Experimental Setup}
\label{sec:ex_protocl}

\noindent
\textbf{Datasets.} There are a few publicly available datasets for offline HSV Systems. In this work, we adopt datasets used in related works \cite{Souza2018_WICNN,Talles2023_AmultitaskApproach4ContrastiveLearning} summarized in Table~\ref{tab:datasets}. 
\\

\input{tables/tab_databases.tex}

\noindent
\textbf{Preprocessing.} As \textit{SigNet-S} is adopted as the backbone for feature extraction, to ensure the reported performance of the model, we have adhered to the same initial preprocessing steps as described in \cite{Hafemann2017_learning} and \cite{Talles2023_AmultitaskApproach4ContrastiveLearning}. First, images are centered on a large canvas, with dimensions determined by the maximum size encountered within each dataset. Next, the image's background is removed using Otsu's algorithm \cite{Otsu1979}, which transforms background pixels to white and foreground pixels to grayscale. Subsequently, the image is inverted to set the background to zero value. Then, all images are resized to the size of 170 pixels in height and 242 pixels in width, and finally, a center crop of size 150x220 is taken. 
 
\noindent \textbf{Classifier.} For classifier comparison in batch context, we employ soft margin SVM with Radial Basis Function (RBF) kernel, following the
experimental protocol defined in \cite{Hafemann2017_learning,Souza2018_WICNN,Talles2023_AmultitaskApproach4ContrastiveLearning}. The SVM regularization parameter is given by $1.0$, and the RBF kernel coefficient hyper-parameter is given by $2^{-11}$. 

\noindent \textbf{Types of signatures.} In this work, there are three types of signatures: genuine (G), which belongs to the claimed user; random forgery (RF), which is a genuine signature that belongs to a user different from the claimed one; and skilled forgery (SK), which belongs to the claimed user but it was produced by a forger. The set of all genuine signatures in a dataset is given by $\mathbb{S}_{G}  = \{\mathbb{S}^{1}_{G},\mathbb{S}^{2}_{G},..., \mathbb{S}^{N}_{G} \}$, where $N$ is the number of users and $\mathbb{S}^{i}_{G}$ refers to the set of genuine signatures of user $i$. Specifically, $\mathbb{S}^{i}_{G} = \{S^{i,1}_{G},S^{i,2}_{G} ,..., S^{i,K}_{G} \}$, where $K$ denotes the number of user's signatures. Likewise, there are sets of random forgeries $\mathbb{S}_{RF}$ and skilled forgery $\mathbb{S}_{SK}$ signatures for each user. 

\noindent \textbf{Data segmentation and generation.} Following \cite{Souza2018_WICNN,Talles2022_ContrastiveLearning}, datasets are split into two disjoint subsets
of users: the development $\mathcal{D}$, employed for training, and the exploitation set $\mathcal{E}$, employed for testing models as shown in Figure~\ref{fig:data_segmentation}. Considering a dichotomy transformation $DT(,)$, a representation model $\phi(\cdot)$, the sets are generated as follows: 

\input{figures/fig_data_segmentation/main}

\begin{itemize}[topsep=0pt, partopsep=0pt, parsep=0pt, itemsep=5pt]
\item \textbf{Development set $\mathcal{D}$}: 
    For each user $i$ in $\mathcal{D}$, \(n\mathcal{D}_G\) genuine signatures are randomly selected forming the set $\mathbb{S}^i_{G_{\mathcal{D}}}$. The genuine signatures in $\mathbb{S}^i_{G_{\mathcal{D}}}$ are paired to form dissimilarity vectors of \textit{positive} class as defined in Equation~\ref{eq:D_set_pos}:
    \begin{itemize}
        \item[] \textbf{Positive set:} 
            \begin{equation}\label{eq:D_set_pos}
                \mathbb{\dot{X}}^{i}_{\mathcal{D}+} = \bigcup\limits_{k=1}^{n\mathcal{D}_G-1}\bigcup\limits_{j=k+1}^{n\mathcal{D}_G} DT(\phi(S_G^{i,k}),\phi(S_G^{i,j}))
            \end{equation}
            with  $S^{i,*}_{G} \in \mathbb{S}^i_{G_\mathcal{D}}$.
    \end{itemize}
    
Additionally, For each user $i$ in $\mathcal{D}$, \(n\mathcal{D}_G / 2\) random forgeries signatures are randomly selected forming the set $\mathbb{S}^i_{{RF}_{\mathcal{D}}}$. Then, \(n\mathcal{D}_G - 1\) genuine signatures in $\mathbb{S}^i_{G_{\mathcal{D}}}$ are paired with all random signatures in $\mathbb{S}^i_{{RF}_{\mathcal{D}}}$, resulting in dissimilarity vectors of \textit{negative} class as defined in Equation~\ref{eq:D_set_negative}:

\begin{itemize}
  \item[]\textbf{Negative set:}
            \begin{equation}\label{eq:D_set_negative}
            \mathbb{\dot{X}}^{i}_{\mathcal{D}-} = \bigcup\limits_{k=1}^{n\mathcal{D}_G-1} \bigcup\limits_{j=1}^{n\mathcal{D}_G/2} DT(\phi(S_G^{i,k}),\phi(S_{RF}^{i,j}))
            \end{equation} 
            with  $S^{i,k}_{G} \in \mathbb{S}^i_{G_\mathcal{D}}$ and $S^{i,j}_{RF} \in \mathbb{S}^i_{RF_\mathcal{D}}$.
\end{itemize}

The final set is formed by the union of $\mathbb{\dot{X}}^{i}_{\mathcal{D}+}$ and $\mathbb{\dot{X}}^{i}_{\mathcal{D}-}$ for all users, consisting of an equal number of positive (\(+\)) and negative (\(-\)) dissimilarity samples from \(n\mathcal{D}\) users. 

\item \textbf{Exploitation set $\mathcal{E}$:} 
For each user $i$ in $\mathcal{E}$, $n\mathcal{E}_R$ reference (genuine) signatures are randomly selected, forming the set $\mathbb{S}^{i}_{R}$. Then, $n\mathcal{E}_C$ signatures of each type (G, RF, SK) are randomly selected resulting in the claimed set $\mathbb{S}^{i}_{C}$ formed by the union of $\mathbb{S}^{i}_{G_\mathcal{E}}$, $\mathbb{S}^{i}_{RF_\mathcal{E}}$, and $\mathbb{S}^{i}_{SK}$ sets of signatures. After that, dissimilarities between all samples in $\mathbb{S}^{i}_{R}$ and $\mathbb{S}^{i}_{C}$ are computed, and the result comprises the exploitation set. This process is defined in Equation~\ref{eq:E_set_user} and \ref{eq:E_set}:

\begin{minipage}{0.5\textwidth}
\begin{equation}\label{eq:E_set_user}
\mathbb{\dot{X}}^{i}_{\mathcal{E}} = \bigcup\limits_{k=1}^{n\mathcal{E}_R} \bigcup\limits_{j=1}^{n\mathcal{E}_C} \{
        \mathbf{\dot{x}}^{i,k,j}_G, \mathbf{\dot{x}}^{i,k,j}_{RF},
        \mathbf{\dot{x}}^{i,k,j}_{SK}
        \}
\end{equation}
\end{minipage}\hfill
\begin{minipage}{0.30\textwidth}
\begin{equation}\label{eq:E_set}
\mathbb{\dot{X}}_{\mathcal{E}} = \bigcup\limits_{i=1}^{n\mathcal{E}} \mathbb{\dot{X}}^{i}_{\mathcal{E}}
\end{equation}
\end{minipage}
\end{itemize}

\begin{itemize}[topsep=0pt, partopsep=0pt, parsep=0pt, itemsep=5pt]
    \item[]Where:
    \item[$\bullet$] $\mathbf{\dot{x}}^{i,k,j}_G = DT(\phi(S^{i,k}_R),\phi(S^{i,j}_G))$  \textit{positive} dissimilarity vector (genuine) 
    \item[$\bullet$] $\mathbf{\dot{x}}^{i,k,j}_{RF} = DT(\phi(S^{i,k}_R),\phi(S^{i,j}_{RF}))$  \textit{negative} dissimilarity vector (random forgery) 
    \item[$\bullet$] $\mathbf{\dot{x}}^{i,k,j}_{SK} = DT(\phi(S^{i,k}_R),\phi(S^{i,j}_{SK}))$ \textit{negative} dissimilarity vector (skilled forgery)
\end{itemize}

\vspace{1pt}

For the present work, dataset segmentation is shown in Tables~\ref{tab:development_set} and~\ref{tab:exploitation_set}.

\input{tables/tab_dev_exp}

\noindent \textbf{Stream generation.} In order to provide a comparable evaluation between batch and stream settings, the batch data generation previously described is extended to an equivalent stream configuration. To generate the stream, the exploitation set $\mathcal{E}$ is converted into a timeline where the whole stream of samples is given by $\mathbb{T} = \{\mathbb{T}_1, \mathbb{T}_2, ..., \mathbb{T}_{\infty}\}$, where each $\mathbb{T}_j$ is a set of arriving signatures as shown in Figure~\ref{fig:stream_T_sig}.

\input{figures/fig_streamT_sig/main}

From the exploitation set $\mathcal{E}$, the set of claimed signatures \(\mathbb{S}_{C}\) is transformed in a stream comprised of  \( n\mathcal{E}_C \) chunks. Each chunk $\mathbb{T}_j$ contains all users in $\mathcal{E}$ requesting the verification of three samples: a genuine, a random forgery, and a skilled forgery signature. 

After generating stream $\mathbb{T}$, it is employed as input to the SHSV system (Figure~\ref{fig:fig_onlineASV}). First, each signature has its features extracted and passed to the stream dichotomy transformation. For this step, the set of reference signatures $\mathbb{S}_R$ from the exploitation set $\mathcal{E}$ is retrieved, then features are extracted, and $DT(,)$ is applied on each corresponding pair of feature vectors. That is, the stream $\mathbb{T}$ results in a stream of dissimilarity sets as defined in Equations~\ref{eq:Tk} and ~\ref{eq:T}:

\begin{minipage}{0.5\textwidth}
\begin{equation}\label{eq:Tk}
    \mathbb{\dot{T}}_j = \bigcup\limits_{i=1}^{n\mathcal{E}} \{ \mathbb{\dot{X}}^{i,j}_G, \mathbb{\dot{X}}^{i,j}_{RF}, \mathbb{\dot{X}}^{i,j}_{SF}\}
\end{equation}
\end{minipage}\hfill
\begin{minipage}{0.30\textwidth}
 \begin{equation}\label{eq:T}
    \mathbb{\dot{T}} = \bigcup\limits_{j=1}^{n\mathcal{E}_C} \mathbb{\dot{T}}_j
\end{equation}
\end{minipage}

\begin{itemize}
    \item[] Where: 
    \item[$\bullet$] $\mathbb{\dot{X}}^{i,j}_G = DT(\phi(\mathbb{S}^i_{R}),\phi(S^{i,j}_G))$ set of \textit{positive} dissimilarity vectors (genuine) 
    \item[$\bullet$] $\mathbb{\dot{X}}^{i,j}_{RF} = DT(\phi(\mathbb{S}^i_{R}),\phi(S^{i,j}_{RF}))$ set of \textit{negative} dissimilarity vectors (random)
    \item[$\bullet$] $\mathbb{\dot{X}}^{i,j}_{SF} = DT(\phi(\mathbb{S}^i_{R}),\phi(S^{i,j}_{SF}))$ set of \textit{negative} dissimilarity vectors (skilled)
\end{itemize}

Stream $\mathbb{\dot{T}}$ is then sent to the WI-classifier for testing, and a decision using the fusion function $\mathtt{g}(\cdot)$ is performed for each signature request. For the present work, stream configuration regarding datasets is shown in Table~\ref{tab:exploitation_set}.

\noindent \textbf{Model initialization.} Models are initialized using all $\mathcal{D}$ sets in Table~\ref{tab:development_set}. 

\noindent \textbf{Model update and evaluation.} In this work, we employ the prequential evaluation approach (also known as test-then-train), the most common method for evaluating data streams~\cite{Haug_2022_Standardized_Evaluation_Data_Streams}. This approach involves continuously testing a predictive model with new arriving samples and then using those same samples to update the model. We assume all instance labels are available after testing and do not employ skilled forgeries for classifier updating.

Given a stream of arriving signatures, $\mathbb{T}$, we define three hyperparameters for model update and evaluation:
\begin{itemize}
    \item \textbf{Chunk size ($c_{size}$)}: 
    The number of signatures the system waits for before updating the model.
    \item \textbf{Window size ($w_{size}$)}: The number of most recently tested signatures used to compute performance metrics.
    \item \textbf{Window step ($w_{step}$)}: The frequency (number of new signatures) at which the metrics are assessed.
\end{itemize}

The stream evaluation employed in this paper is summarized in Table~\ref{tab:stream_evaluation}. While the entire block $\mathbb{T}_k$ is tested, only the genuine and random forgery signatures are sent to update the classifier. A window smaller than the chunk size is employed to observe the evolution more frequently. 

\input{tables/tab_stream_eval}

With the test results, the Equal Error Rate (EER) using a global threshold is measured at every window. The experiments are repeated five times, and the average results and standard deviation are computed.

%% file: tables/tab_databases.tex
\vspace{-2em}
\begin{table}
\caption{Commonly used datasets for Offline Signature Verification}
\label{tab:datasets}
\centering

\begin{tabular}{lllccc}
\hline
Ref & Dataset Name & Language & Users & Genuine signatures & Forgeries \\
\hline

\cite{Kalera2004_DB_CEDAR}	&	CEDAR	&	Western	&	55	&		24	&	24	\\
\cite{Ferrer2017_DB_unifiedSynthesizer}	&	GPDS Synthetic	&	Western	&	10000	&		24	&	30	\\
\cite{OrtegaGarcia2003_DB_MCYT}	&	MCYT-75	&	Western	&	75	&		15	&	15	\\

\hline
\end{tabular}
\end{table}

%% file: figures/fig_data_segmentation/main.tex
\begin{figure*}
    \centering
    \includegraphics[trim=2cm 0.4cm 9cm 0cm, width=0.5\linewidth]{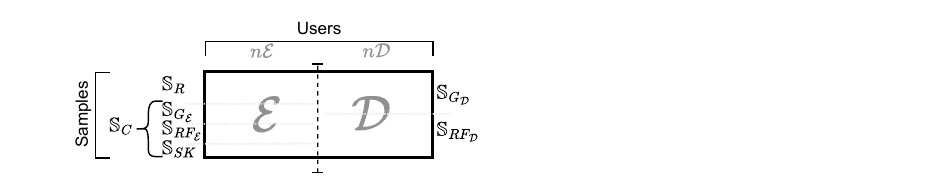}
    \caption{Data segmentation into development $\mathcal{D}$ and exploitation $\mathcal{E}$ sets. To generate $\mathcal{E}$, a set of references $\mathbb{S}_R$ and claimed signatures $\mathbb{S}_C$ are randomly selected for all $n\mathcal{E}$ users. $\mathbb{S}_C$ contains genuine, random forgery, and skilled forgery samples. To generate $\mathcal{D}$, a set of genuine $\mathbb{S}_{G_\mathcal{D}}$ and random forgery $\mathbb{S}_{RF_\mathcal{D}}$ are randomly chosen for all $n\mathcal{D}$ users. Selected samples are utilized to perform dissimilarity transformations as defined in Equations~\ref{eq:D_set_pos},~\ref{eq:D_set_negative},~\ref{eq:E_set_user}, and~\ref{eq:E_set}.}
    \label{fig:data_segmentation}
\end{figure*}

%% file: tables/tab_dev_exp.tex
\begin{table}[h]
\centering
\caption{Data segmentation for each dataset}
\label{tab:sets}

\begin{subtable}{\textwidth}
\centering
\caption{Development set ($\mathcal{D}$).}
\label{tab:development_set}
\begin{tabular}{
l@{\hskip 3pt}
l@{\hskip 2pt}
c@{\hskip 1pt}
c@{\hskip 1pt}
l@{\hskip 1pt}
l@{\hskip 5pt}
l@{\hskip 0pt}
c}
\hline
\textbf{Data}
&\multicolumn{1}{c}{\textbf{\#Users}}
&
&\textbf{\#G sig.}
&
&\multicolumn{1}{c}{\textbf{Neg. class}}
&\multicolumn{1}{c}{\textbf{Pos. class}}
&\multicolumn{1}{c}{\textbf{\#Samples}}
\\
\textbf{}
&\multicolumn{1}{c}{\textbf{($n\mathcal{D}$)}}
&\textbf{}
&\textbf{($n\mathcal{D}_G$)}
&
& \multicolumn{2}{c}{Dissimilarity between:}
&
 \\ \cline{1-2} \cline{4-8}
GPDS-S
&\{5, 10, 50, 581 (all)\}
&$\times$
&\{2, 6, 12\}
&
& 11G \& 6RF
& 12G of each user
& $n\mathcal{D} \cdot 66 \cdot 2$
\\ 
\\
\hline
\end{tabular}
\end{subtable}

\vspace{0pt}

\begin{subtable}{\textwidth}
\centering
\caption{Exploitation set ($\mathcal{E}$).}
\label{tab:exploitation_set}
\begin{tabular}{
c@{\hskip 1pt} 
c@{\hskip 0pt} 
c@{\hskip 10pt} 
c@{\hskip 1pt} 
c@{\hskip 10pt} 
c@{\hskip 5pt}
c@{\hskip 10pt}
c
}
\hline
\multicolumn{1}{c}{\textbf{Data}}
&\multicolumn{1}{c}{\textbf{\#Users}}
&\multicolumn{1}{c}{\textbf{\#Ref. signatures}}
&\multicolumn{2}{c}{\textbf{Claimed signatures}}

&
&\multicolumn{2}{c}{\textbf{Stream $\mathbb{T}$}}

\\
\textbf{}
&\multicolumn{1}{c}{\textbf{($n\mathcal{E}$)}}
&\multicolumn{1}{c}{\textbf{($n\mathcal{E}_R$)}}
&\multicolumn{1}{c}{\textbf{\#($n\mathcal{E}_C$)}}
&Set
&
&$\mathbb{T}_j$ size
&$j$ value
 \\ \cline{1-5} \cline{7-8}
GPDS-S
&300
&\{1, 2, 3, 5, 10, 12\}
&10
& 10G, 10RF, 10SK
&
&\multirow{3}{*}{$n\mathcal{E} \cdot 3 $}
& \multirow{3}{*}{$n\mathcal{E}_C$}
 \\
CEDAR
&55
&\{10\}
&10
& 10G, 10RF, 10SK
&
&
&
 \\
MCYT
&75
&\{10\}
&5
& 5G, 5RF, 5SK
&
&
&
 \\
\hline
\end{tabular}
\end{subtable}

\end{table}

%% file: figures/fig_streamT_sig/main.tex
\begin{figure*}
    \centering
    \includegraphics[trim=1.1cm 0cm 4.5cm 0cm, width=\linewidth]{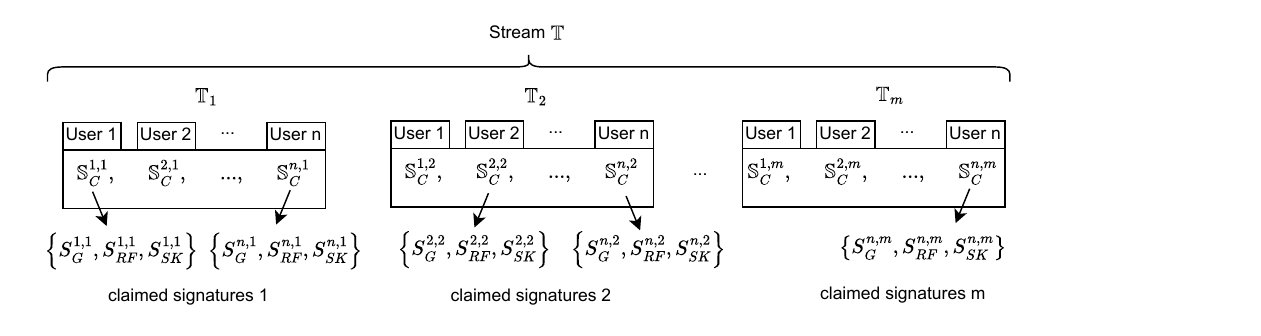}
    \caption{Stream $\mathbb{T}$ of claimed signatures obtained from the exploitation set $\mathcal{E}$. The number of users $n\mathcal{E}$ is represented by $n$, while the number of claimed signatures $n\mathcal{E}_C$ is denoted by $m$. $\mathbb{S}^{i,j}_C$ is the  $j$-th set of claimed signature of user $i$, where $\mathbb{S}^{i,j}_C = \{S^{i,j}_G,S^{i,j}_{RF},S^{i,j}_{SK}\}$, with $S^{i,j}_G \in \mathbb{S}^i_{G_\mathcal{E}}$, $S^{i,j}_{RF} \in \mathbb{S}^i_{RF_\mathcal{E}}$ and $S^{i,j}_{SK} \in \mathbb{S}^i_{SK}$. G: genuine, RF: random forgery, SK: skilled forgery signature.}
    \label{fig:stream_T_sig}
\end{figure*}

%% file: tables/tab_stream_eval.tex
\vspace{-1em}
\begin{table}
\caption{Stream evaluation.}
\label{tab:stream_evaluation}
\centering
\begin{tabular}{
c@{\hskip 10pt} 
c@{\hskip 15pt} 
c@{\hskip 15pt} 
c 
}
\hline
\multirow{2}{*}{Stream} & \multicolumn{2}{c}{Chunk size} & \multirow{2}{*}{Window size and step} \\
\cline{2-3}
 & Test & Training & \\
\hline
GPDS-S & 900 ($n\mathcal{E} \cdot 3$) & 600 ($n\mathcal{E} \cdot 2$) & 400 \\
CEDAR + MCYT & 300 & 200 & 200 \\
\cline{1-4}
Type of sig.: & G, RF, and SK & G and RF & G and SK \\
\hline
\end{tabular}
\end{table}

%% file: sections/4.2_experimental_results.tex
\section{Experimental Results}
\label{sec:ex_results}

\input{figures/fig_users_vs_sig/main}

\noindent \textbf{Batch evaluation.}
Figure~\ref{fig:fig_users_vs_sig} presents results for batch-trained SVM and SGD considering different number of users $n\mathcal{D}$ ($\#U$) and genuine signatures $n\mathcal{D}_G$ ($\#G$) (Table~\ref{tab:development_set}). Interestingly, increasing the number of training samples does not necessarily guarantee improved performance. For example, the configuration with U=50 users and G=2 genuine signatures per user (orange dotted line) during training, resulting in 100 samples (S:100), achieves better results than the configuration with U=10 users and G=12 signatures (blue solid line, S:1320), despite having fewer signatures overall. This suggests that the number of users available during the development phase plays a crucial role in achieving system generalization. Conversely, when the number of users remains constant, increasing the number of samples per user during training leads to improved performance. 

\input{figures/fig_stream_sgd_svm/main}

Furthermore, SGD exhibits greater sensitivity to limited initial data than SVM, although both achieve comparable performance when trained with all users and samples. Overall, the results indicate a trend of decreasing error rates with an increase in the number of reference signatures. Please refer to Table 1 in the supplementary material for a comprehensive set of results.

\noindent \textbf{Stream evaluation.}
Figure~\ref{fig:fig_stream_sgd_svm} presents the performance comparison between the SVM and the proposed SHSV method when the exploitation set is transformed into a continuous stream of incoming signatures. At a certain point, SHSV surpasses SVM for all initial training configurations, being more pronounced when there is a limited number of users and signatures for pre-training models. Recently, \cite{Talles2023_AmultitaskApproach4ContrastiveLearning} reported an EER of $7.93 \pm 0.30$ for a writer-independent approach using global thresholds on GPDS Synthetic with 12 reference signatures. In their study, the authors utilized \textit{SigNet-S} as the feature extractor, selected 2000 users for training an SVM classifier, and conducted tests on the GPDS-S-300 dataset. In contrast, SHSV achieves comparable results while requiring significantly fewer users for initial training. This finding highlights the effectiveness of SHSV in real-world scenarios with restricted sample availability. Moreover, SHSV accommodates the dynamic nature of handwriting signatures by enabling a continuous adaptation of the system, leading to improved performance over time.

\input{tables/tab_stream_svm_sgd_last}

Table~\ref{tab:stream_eer_last_t} presents the performance of SVM and SHSV on the final chunk of the signature stream, evaluated across different numbers of reference signatures used for the fusion function. Consistent with the batch setting results, performance improves with an increasing number of stored signatures per user. Findings also highlight the discrepancy between SVM and SHSV performance, which becomes more pronounced as the number of users in the training phase decreases. Unlike SVM, SHSV consistently exhibits improved performance over time, achieving better or comparable results regardless of the initial development configuration.

\input{figures/fig_stream_sgd_svm_cedar_mcyt_mixed/main}

SHSV is particularly interesting for handling signatures from unknown distributions due to its ability to learn over time. Figure~\ref{fig:stream_svm_sgd_r10_ch200_g12_tr_sgpds_signets_ts_cedar_mcyt_mixed_skilled_global} shows the performance of models pre-trained on GPDS Synthetic data when they receive signatures coming randomly from CEDAR stream and MCYT stream. While initially affected by the change, SHSV outperforms SVM, especially when few users are available at the beginning. Please see Table 2 in the supplementary material for a detailed set of results. 

In summary, the proposed SHSV system consistently demonstrates superior performance and adaptability compared to the traditional SVM approach in various scenarios. Its resilience to limited initial training data and its continuous adaptation capabilities make SHSV particularly well-suited for real-world handwriting signature verification tasks where data availability and signature variability pose significant challenges.

%% file: figures/fig_users_vs_sig/main.tex
\begin{figure}[h]
    \centering

    \vspace{0.1cm} 
    \begin{subfigure}{0.49\textwidth}
        \includegraphics[trim=0 200 0 200, clip,width=\linewidth]{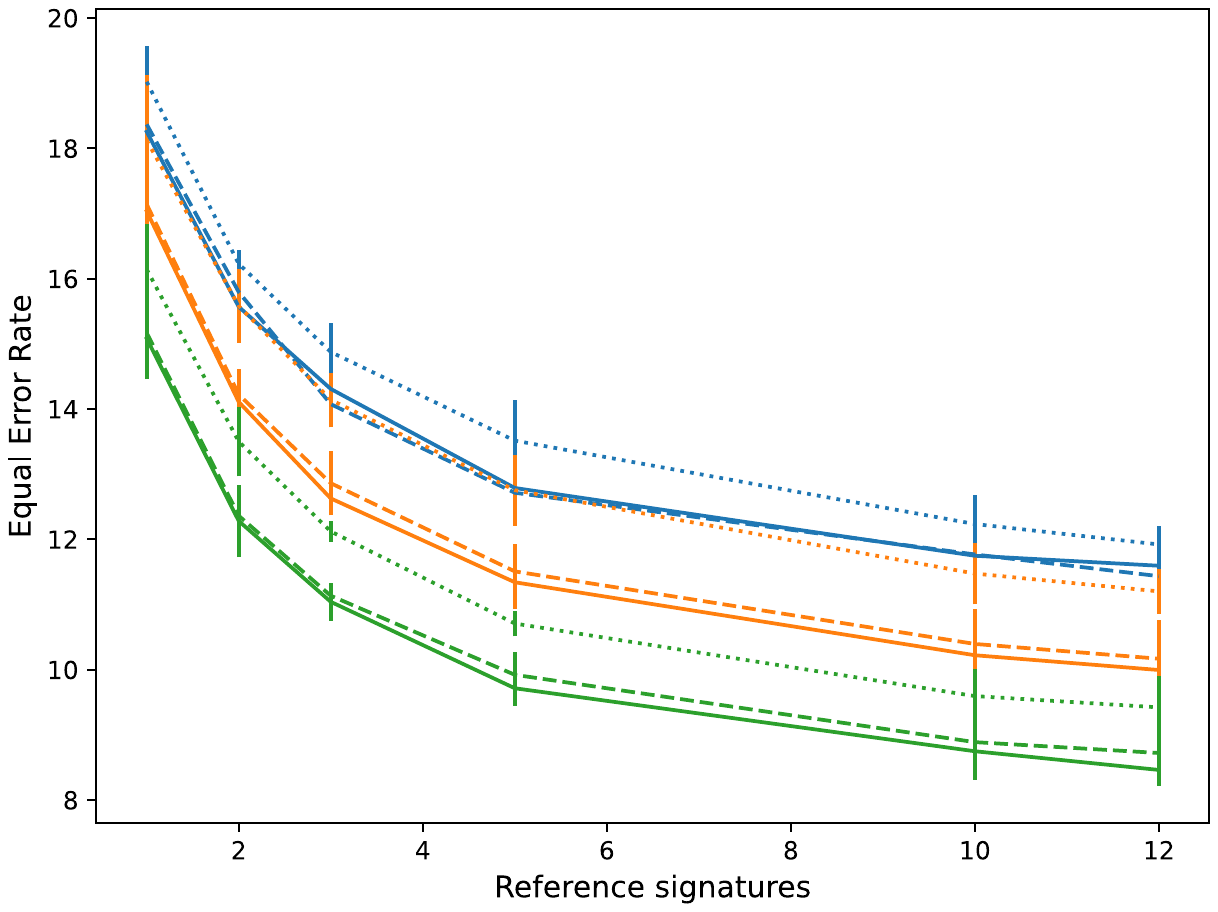}
        \caption{SVM}
        \label{fig:fig_users_vs_sig_a}
    \end{subfigure}
    \hfill
    \begin{subfigure}{0.49\textwidth}
        \includegraphics[trim=0 200 0 200, clip,width=\linewidth]{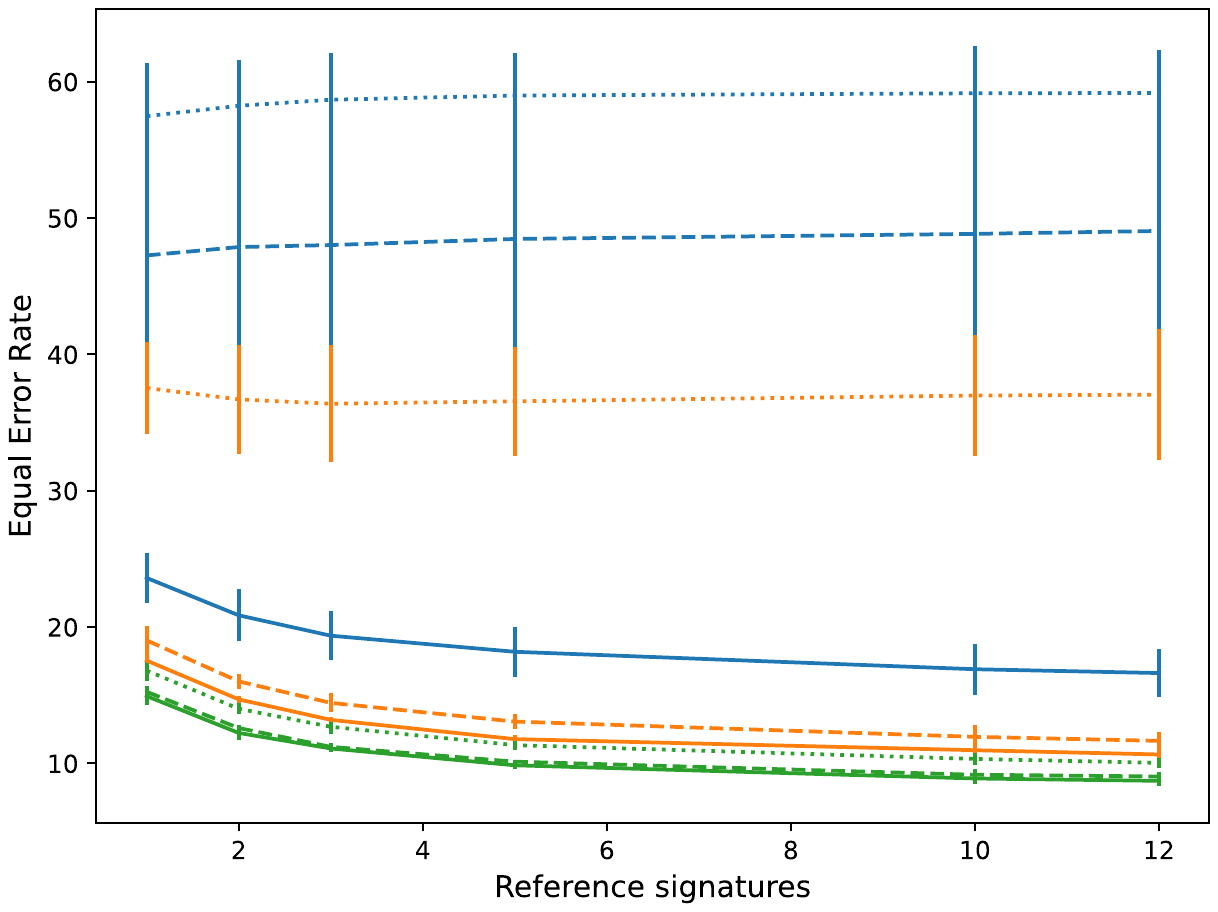}
        \caption{SGD}
        \label{fig:fig_users_vs_sig_b}
    \end{subfigure}

    \vspace{0.1cm} 
    \begin{subfigure}{\textwidth}
        \centering
        \includegraphics[trim=0 165 0 610, clip,width=0.8\linewidth]{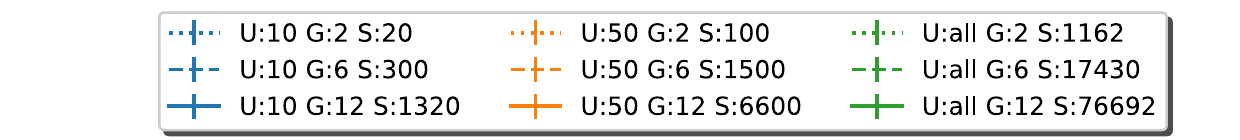}
    \end{subfigure}
    \caption{Skilled forgery detection in batch settings for different development sets. \#U and \#G denote, respectively, the number of users $n\mathcal{D}$ and genuine signatures $n\mathcal{D}_G$ employed during training. \#S denotes the resulting number of samples.}
    \label{fig:fig_users_vs_sig}
\end{figure}

%% file: figures/fig_stream_sgd_svm/main.tex
\begin{figure}[!]
    \centering
    
    \begin{subfigure}{0.49\textwidth}
        \includegraphics[trim=0 200 0 200, clip,width=\linewidth]{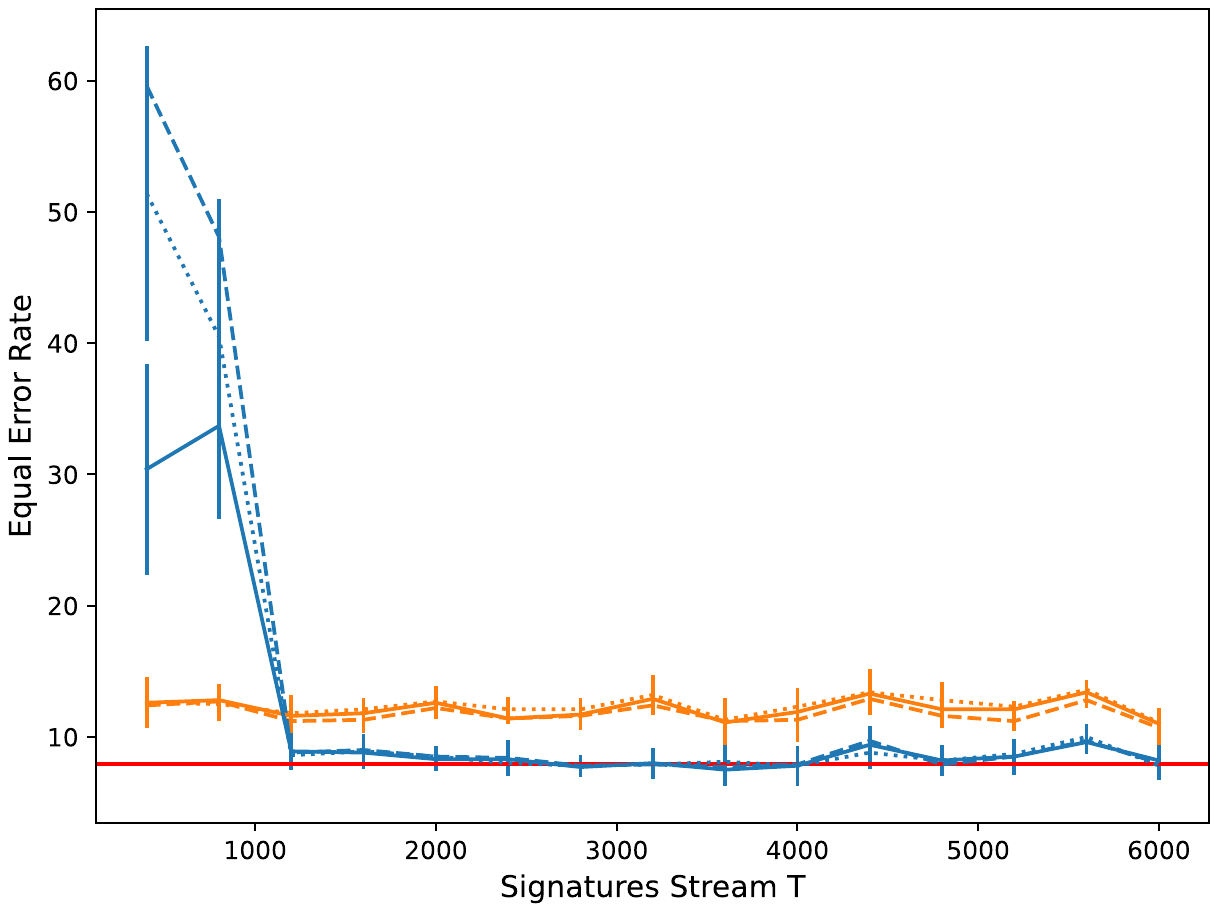}
        \caption{\#Users ($n\mathcal{D}$) = 5}
        
    \end{subfigure}
    \hfill
    \begin{subfigure}{0.49\textwidth}
        \includegraphics[trim=0 200 0 200, clip,width=\linewidth]{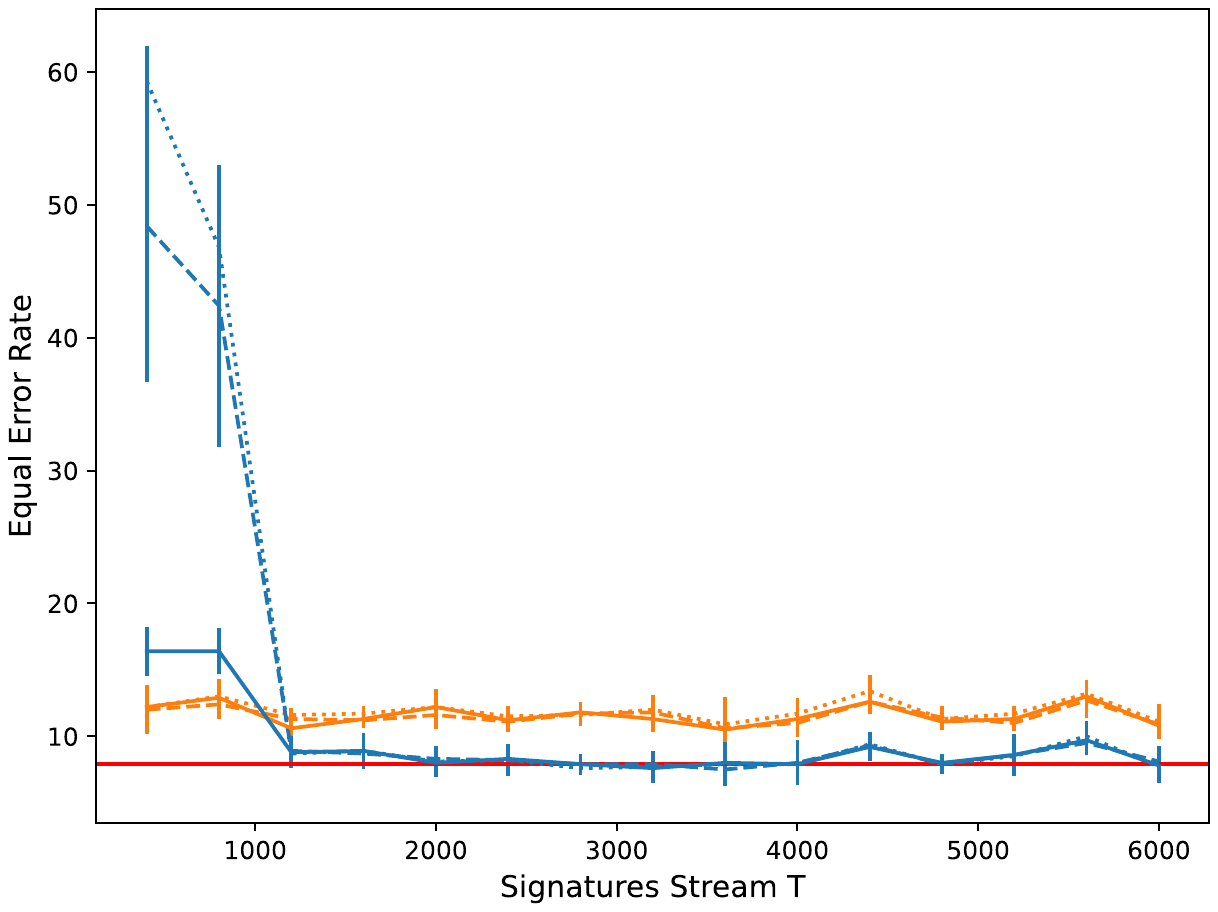}
        \caption{\#Users ($n\mathcal{D}$) = 10}
    
    \end{subfigure}
    
    \begin{subfigure}{0.49\textwidth}
        \includegraphics[trim=0 200 0 200, clip,width=\linewidth]{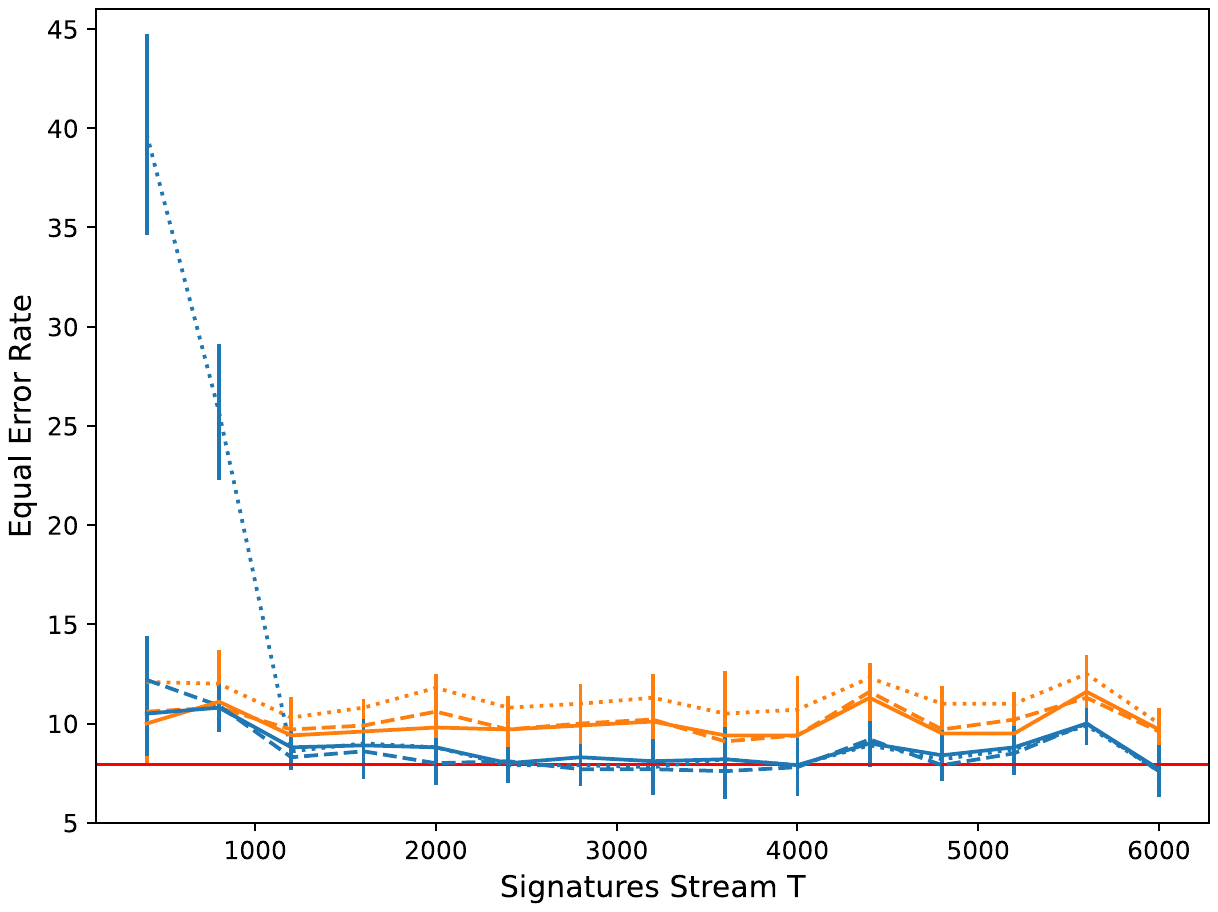}
        \caption{\#Users ($n\mathcal{D}$) = 50}
    \end{subfigure}
    \hfill
    \begin{subfigure}{0.49\textwidth}
        \includegraphics[trim=0 200 0 200, clip,width=\linewidth]{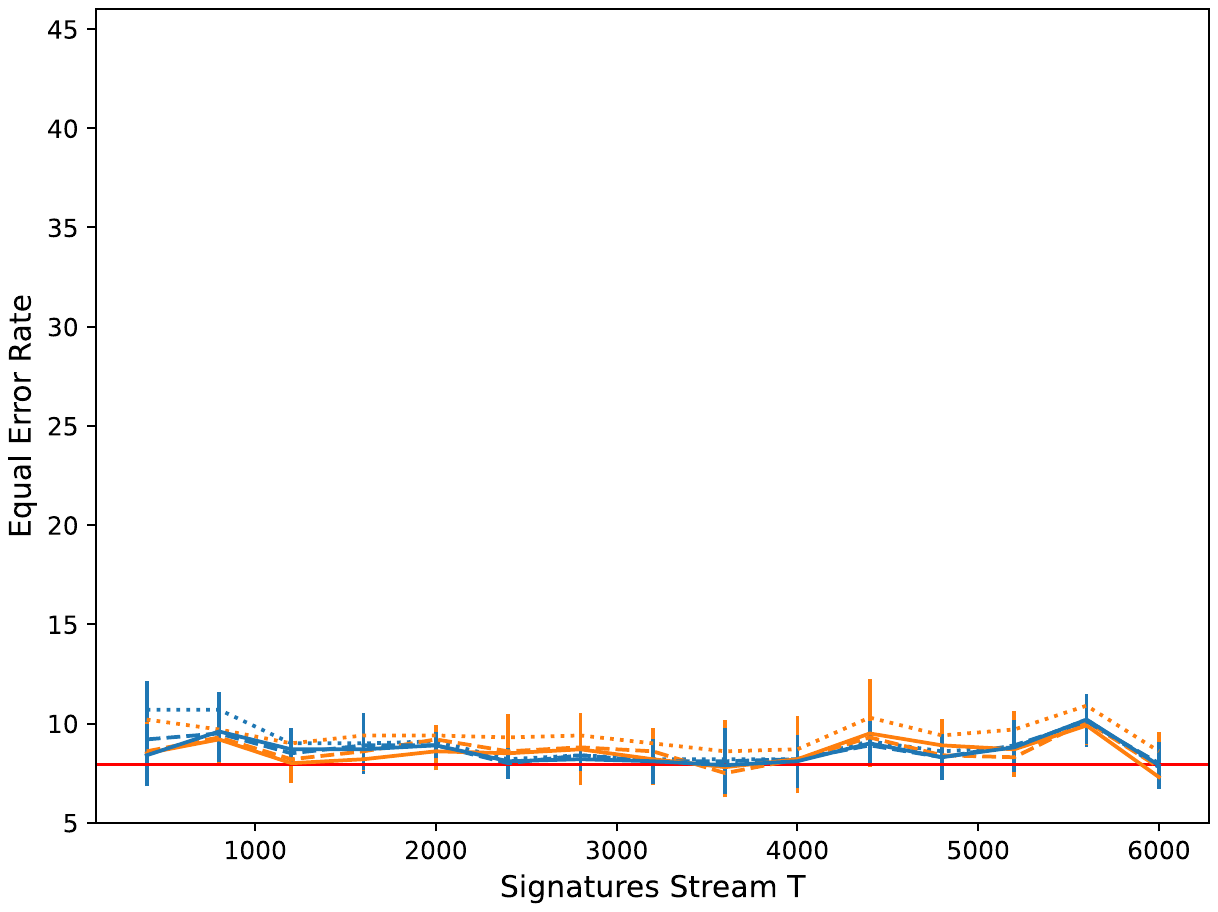}
        \caption{\#Users ($n\mathcal{D}$) = All}
        
    \end{subfigure}
    \vspace{0.1cm} 
    \begin{subfigure}{\textwidth}
        \centering
        \includegraphics[trim=0 165 0 620, clip,width=\linewidth]{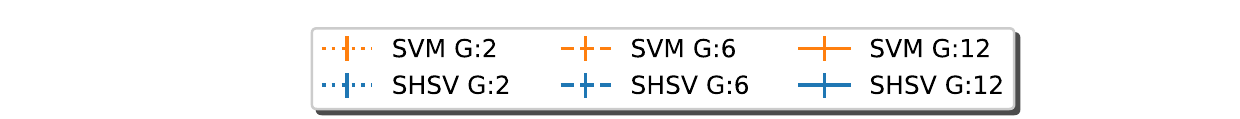}

    \end{subfigure}
    
    \caption{Stream evaluation of skilled forgery detection on GPDS Synthetic using SVM and the Stream HSV (SHSV). The evaluation considers 12 reference signatures with a Max fusion for decision-making. SHSV is updated after every training chunk, employing only genuine and random forgery signatures, and evaluated on every window (Table~\ref{tab:stream_evaluation}). \#Users ($n\mathcal{D}$) and \#G denote the number of users and genuine signatures used in the initial training, respectively. The horizontal red line shows the result ($7.93$) reported in \cite{Talles2023_AmultitaskApproach4ContrastiveLearning} for 12 reference signatures, \#G=12, and $n\mathcal{D}$ = 2000.}
    \label{fig:fig_stream_sgd_svm}
\end{figure}

%% file: tables/tab_stream_svm_sgd_last.tex
\begin{table}[!]
        \caption{EER on the last chunk of signatures stream. Results show skilled forgery detection on GPDS Synthetic using SVM and Stream HSV (SHSV). $n\mathcal{D}$ and $n\mathcal{D}_G$ refer to the initial training setup (Table~\ref{tab:development_set}). $n\mathcal{E}_R$ denotes the number of reference signatures (Table~\ref{tab:exploitation_set}). Stream evaluation performed according to Table~\ref{tab:stream_evaluation}.}
        \label{tab:stream_eer_last_t}
        \centering
                \begin{tabular}{ccccccccc}
                         \cline{1-5}\cline{7-9}
                         \multirow{2}{*}{$n\mathcal{D}$} 
                         &\multirow{2}{*}{$n\mathcal{E}_R$} 
                         & \multicolumn{3}{c}{\#G sig. ($n\mathcal{D}_G$) = 2}     
                         &  
                         & \multicolumn{3}{c}{\#G sig. ($n\mathcal{D}_G$) = 12}    \\ \cline{3-5}\cline{7-9}

                        & & $SVM_{last}$             & $SHSV_{last}$   &$\Delta$  & &$SVM_{last}$          & $SHSV_{last}$ & $\Delta$ \\ \cline{1-5}\cline{7-9}

			\multirow{4}{*}{2} & 1 & 19.20 (2.73) & 14.00 (2.09) & \textcolor{blue}{5.20} &  & 19.00 (2.26) & 15.00 (1.41) & \textcolor{blue}{4.00} \\
			 & 2 & 17.40 (2.22) & 12.00 (2.40) & \textcolor{blue}{5.40} &  & 17.00 (1.12) & 12.90 (1.98) & \textcolor{blue}{4.10} \\
			 & 5 & 14.20 (1.48) & 9.20 (1.20) & \textcolor{blue}{5.00} &  & 14.10 (1.08) & 9.00 (1.27) & \textcolor{blue}{5.10} \\
			 & 12 & 12.10 (0.55) & 7.70 (1.15) & \textcolor{blue}{4.40} &  & 11.30 (0.57) & 7.80 (1.10) & \textcolor{blue}{3.50} \\ \cline{1-5}\cline{7-9}
			\multirow{4}{*}{10} & 1 & 18.50 (2.12) & 13.70 (2.05) & \textcolor{blue}{4.80} &  & 17.70 (2.17) & 13.80 (2.08) & \textcolor{blue}{3.90} \\
			 & 2 & 16.00 (1.41) & 12.20 (2.61) & \textcolor{blue}{3.80} &  & 15.20 (2.05) & 12.10 (2.88) & \textcolor{blue}{3.10} \\
			 & 5 & 13.00 (1.37) & 9.20 (1.35) & \textcolor{blue}{3.80} &  & 11.90 (1.08) & 9.10 (1.19) & \textcolor{blue}{2.80} \\
			 & 12 & 11.10 (1.29) & 7.70 (1.15) & \textcolor{blue}{3.40} &  & 10.80 (1.04) & 7.80 (1.30) & \textcolor{blue}{3.00} \\ \cline{1-5}\cline{7-9}
			\multirow{4}{*}{50} & 1 & 17.90 (1.67) & 14.00 (2.26) & \textcolor{blue}{3.90} &  & 16.60 (1.78) & 14.20 (2.05) & \textcolor{blue}{2.40} \\
			 & 2 & 14.60 (1.14) & 12.20 (1.89) & \textcolor{blue}{2.40} &  & 14.90 (1.39) & 12.10 (1.78) & \textcolor{blue}{2.80} \\
			 & 5 & 12.30 (1.15) & 9.20 (1.15) & \textcolor{blue}{3.10} &  & 11.30 (1.04) & 9.30 (1.30) & \textcolor{blue}{2.00} \\
			 & 12 & 10.00 (0.71) & 7.60 (0.96) & \textcolor{blue}{2.40} &  & 9.70 (0.76) & 7.70 (0.84) & \textcolor{blue}{2.00} \\ \cline{1-5}\cline{7-9}
			\multirow{4}{*}{All} & 1 & 16.00 (1.77) & 14.50 (2.21) & \textcolor{blue}{1.50} &  & 14.40 (2.19) & 13.90 (2.43) & \textcolor{blue}{0.50} \\
			 & 2 & 13.30 (1.57) & 11.90 (1.98) & \textcolor{blue}{1.40} &  & 12.70 (1.25) & 11.80 (1.92) & \textcolor{blue}{0.90} \\
			 & 5 & 10.70 (1.52) & 9.30 (1.30) & \textcolor{blue}{1.40} &  & 9.00 (1.62) & 9.10 (1.52) & \textcolor{red}{0.10} \\
			 & 12 & 8.60 (0.96) & 7.80 (0.84) & \textcolor{blue}{0.80} &  & 7.30 (0.57) & 7.90 (1.19) & \textcolor{red}{0.60} \\

                        \hline
                \end{tabular}
        
\end{table}

%% file: figures/fig_stream_sgd_svm_cedar_mcyt_mixed/main.tex
\begin{figure}[H]
    \centering
    
    
    
    \begin{subfigure}{0.49\textwidth}
        \includegraphics[trim=0 200 0 200, clip,width=\linewidth]{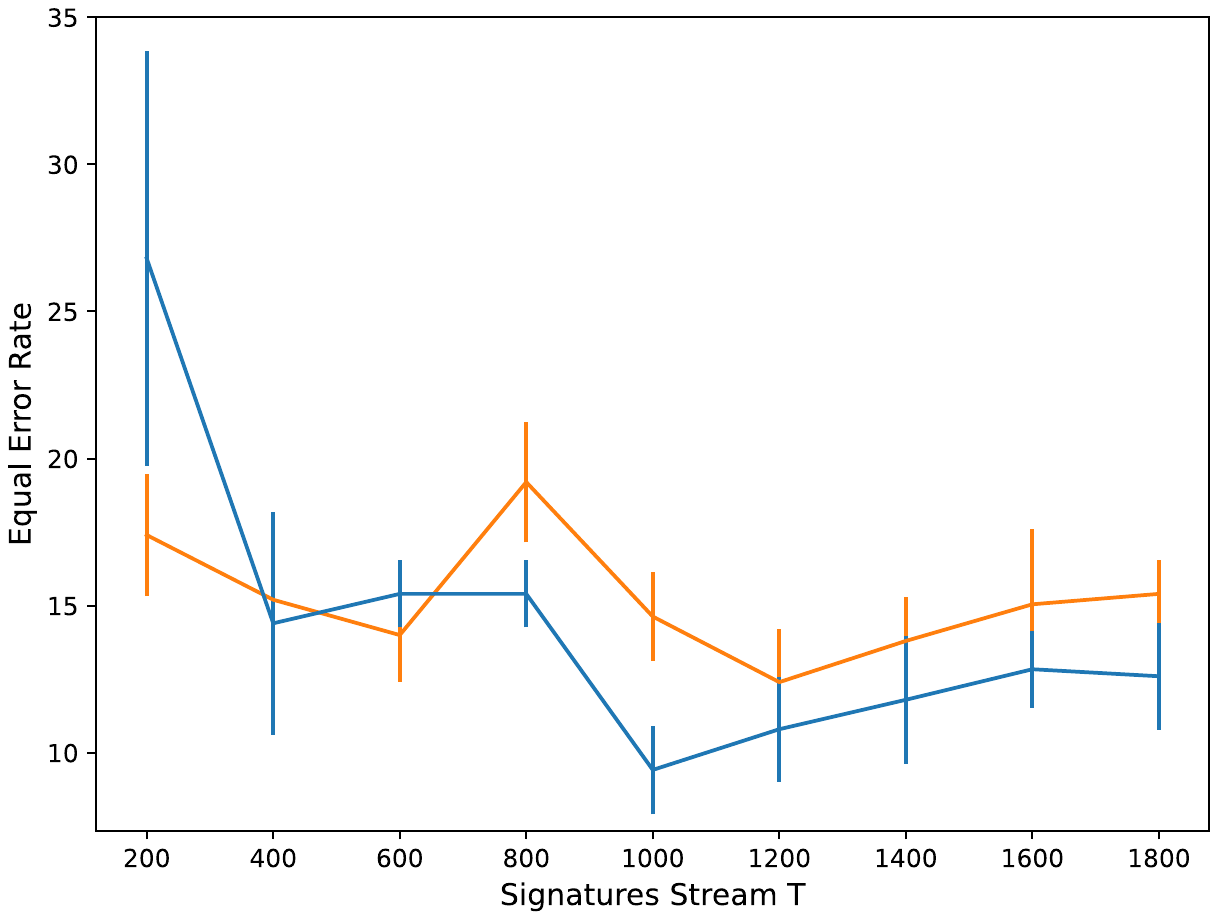}
        \caption{\#Users ($n\mathcal{D}$) = 5, \#G sig = 12}
    \end{subfigure}
    \hfill
    \begin{subfigure}{0.49\textwidth}
        \includegraphics[trim=0 200 0 200, clip,width=\linewidth]{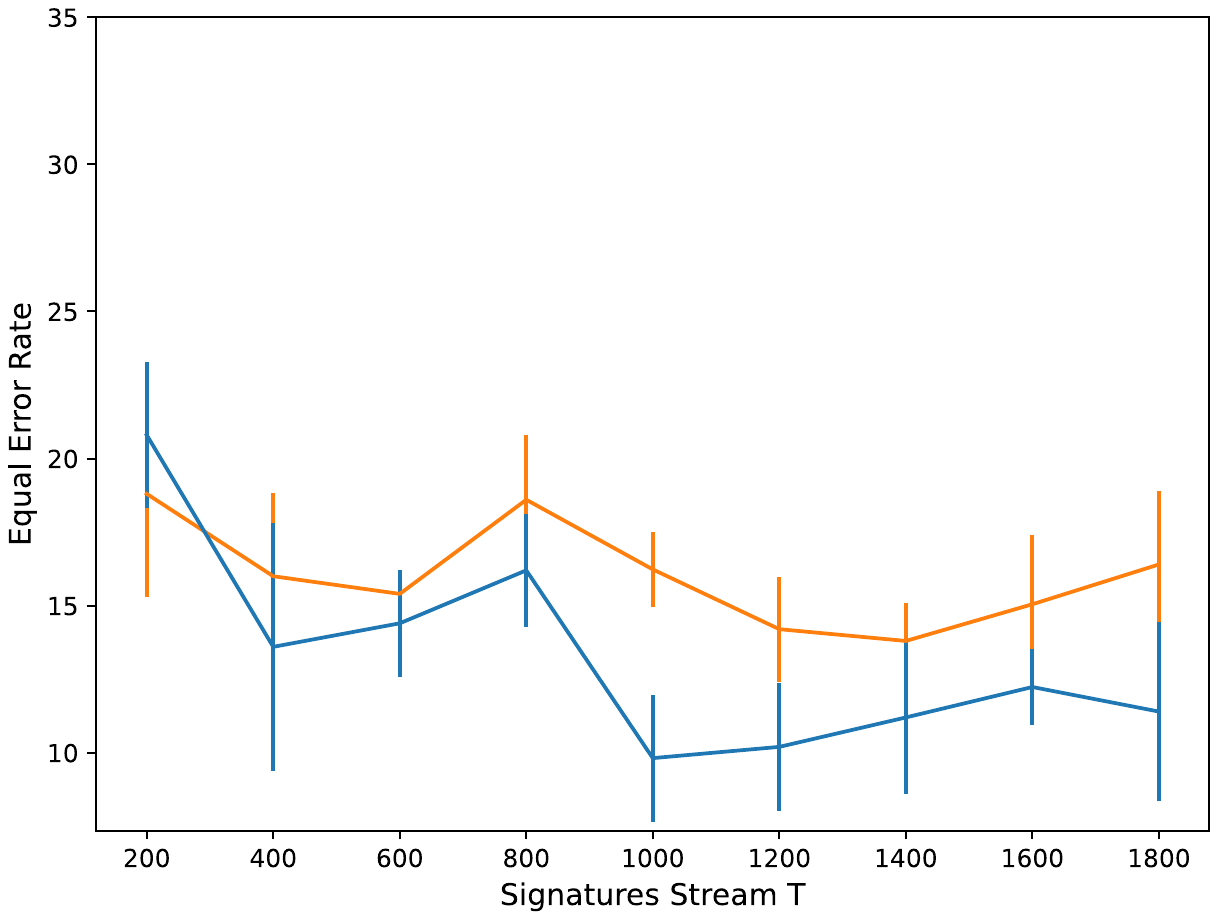}
        \caption{\#Users ($n\mathcal{D}$) = 10, \#G sig = 12}
    \end{subfigure}
    
    \vspace{0.1cm} 
    \begin{subfigure}{0.49\textwidth}
        \includegraphics[trim=0 200 0 200, clip,width=\linewidth]{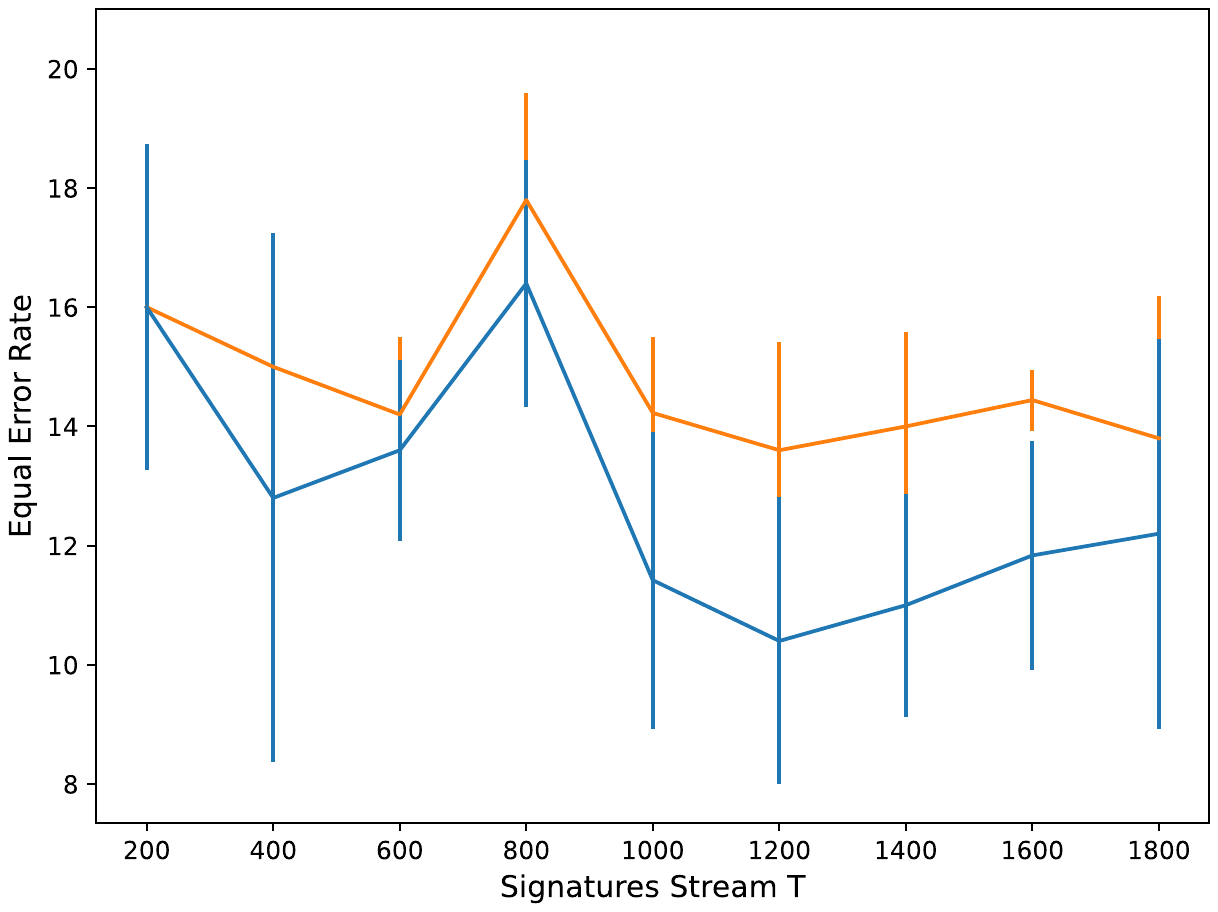}
        \caption{\#Users ($n\mathcal{D}$) = 50, \#G sig = 12}
    \end{subfigure}
    \hfill
    \begin{subfigure}{0.49\textwidth}
        \includegraphics[trim=0 200 0 200, clip,width=\linewidth]{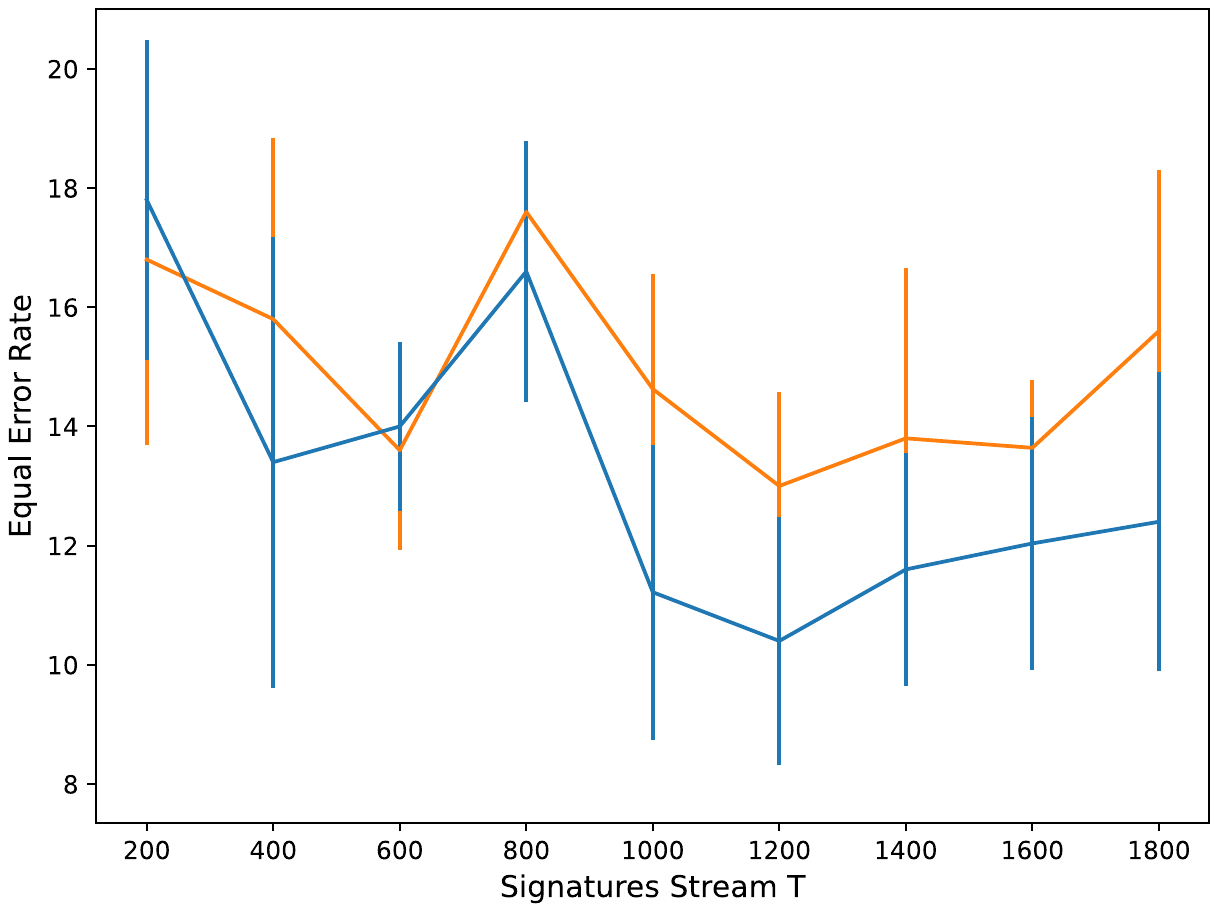}
        \caption{\#Users ($n\mathcal{D}$) = All, \#G sig = 12}
    \end{subfigure}
    
    \vspace{0.1cm} 
    \begin{subfigure}{\textwidth}
        \centering
        \includegraphics[trim=0 160 0 650, clip,width=\linewidth]{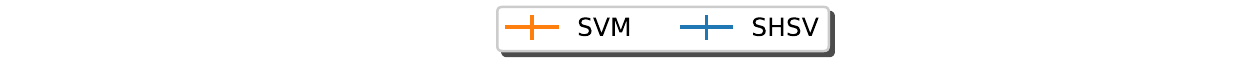}
    \end{subfigure}
    
    \caption{Stream evaluation of skilled forgery detection on signatures randomly coming from the CEDAR and MCYT streams using SVM and the Stream HSV (SHSV). The evaluation considers 10 reference signatures with a Max fusion for decision-making. SHSV is updated after every training chunk and evaluated on every window (Table~\ref{tab:stream_evaluation}). \#Users ($n\mathcal{D}$) and \#G denote the number of users and genuine signatures used in the initial training. }
    \label{fig:stream_svm_sgd_r10_ch200_g12_tr_sgpds_signets_ts_cedar_mcyt_mixed_skilled_global}
\end{figure}

%% file: sections/5_conclusion.tex
\section{Conclusion}
\label{sec:conclusion}

This work proposes a novel handwriting signature verification approach called SHSV. SHSV treats signatures as continuous data streams and updates the system dynamically. To achieve this, we introduce a stream generation approach compatible with standard batch evaluation settings.

Experimental results in batch settings demonstrated that having a high number of users is more crucial than the sheer volume of signatures, indicating an overall improvement in performance when more users are available at initial training. Results also showed that SHSV overcame the problem of limited training data by incorporating new information over time,  demonstrating superior performance compared to the SVM approach across different scenarios under stream configuration.

Future work may include using partially labeled data to explore scenarios where labels are not available for all test samples, as well as investigating the trade-off between adapting the representation model and the WI-classifier over time.